%% file: main.tex
\documentclass[10pt,a4paper]{article}
\usepackage[margin=1in]{geometry}
\usepackage{fontspec}
\usepackage{microtype}
\usepackage{amsmath,amssymb,booktabs,graphicx,array,tabularx}
\usepackage{enumitem,placeins,float,fancyvrb}
\usepackage{subcaption,pgfplots}
\usepackage[numbers,square,sort&compress]{natbib}
\usepackage[hidelinks]{hyperref}
\usetikzlibrary{arrows.meta,positioning,shapes.geometric,fit,calc}
\pgfplotsset{compat=1.18}
\setlist{nosep,leftmargin=*}
\newif\ifanglicomments
\anglicommentsfalse
\ifdefined\hideangli\anglicommentsfalse\fi
\definecolor{anglicolor}{HTML}{9E2A2B}
\DeclareRobustCommand{\angli}[1]{\ifanglicomments{\color{anglicolor}\small\textsf{[Ang Li: #1]}}\fi}
\definecolor{staticcolor}{HTML}{444444}
\definecolor{experiencecolor}{HTML}{0072B2}
\definecolor{failurecolor}{HTML}{D55E00}
\definecolor{bankcolor}{HTML}{009E73}
\definecolor{metacolor}{HTML}{8064A2}
\definecolor{evocolor}{HTML}{C28E00}
\definecolor{rrsicolor}{HTML}{0072B2}
\definecolor{bluephase}{HTML}{397AA8}
\definecolor{orangephase}{HTML}{C75A31}
\pgfplotsset{
  costaxis/.style={
    width=0.98\linewidth,height=4.25cm,
    xlabel={Total tokens (M)},ylabel={Accuracy (\%)},
    tick label style={font=\scriptsize},label style={font=\small},
    grid=major,grid style={gray!18},axis line style={gray!65},
    scaled x ticks=false,scaled y ticks=false,
    /pgf/number format/1000 sep={},clip=false},
  method0/.style={color=staticcolor,mark=*,mark size=2.5pt},
  method1/.style={color=experiencecolor,mark=triangle*,mark size=3pt},
  method2/.style={color=failurecolor,mark=diamond*,mark size=3.5pt},
  method3/.style={color=bankcolor,mark=square*,mark size=2.5pt},
  method4/.style={color=metacolor,mark=pentagon*,mark size=3pt},
  method5/.style={color=evocolor,mark=star,mark size=3pt},
  method6/.style={color=rrsicolor,mark=triangle*,mark size=3pt}}
\newcommand{\costpoint}[6]{%
  \addplot[gray!50,thin,forget plot] coordinates {(#2,#3) (#4,#5)};
  \addplot[only marks,#1,forget plot] coordinates {(#2,#3)};
  \node[font=\scriptsize,fill=white,inner sep=1pt]
    at (axis cs:#4,#5) {#6};}

\newcommand{\notebaseline}{SRN}

\newcommand{\dsmodel}{DeepSeek V4.1 Flash}
\newcommand{\lunamodel}{GPT-6 Luna}
\newcommand{\meanaccgain}{4.7}
\newcommand{\meancostincrease}{3.6}
\newcommand{\meanmetasaving}{65.6}
\newcommand{\meanevosaving}{71.0}
\newcommand{\meanrrsisaving}{68.2}

\title{The RSI Tax:\\When Recursive Self-Improvement Is Not Worth the Cost}
\author{Yitong Zhang$^{1}$ \qquad Ang Li$^{2}$ \qquad Yisen Wang$^{2}$\\[0.6em]
  {\small $^{1}$School of Physics, Peking University} \\
  {\small $^{2}$State Key Lab of General Artificial Intelligence,}\\
  {\small School of Intelligence Science and Technology, Peking University}}
\date{}

\begin{document}
\maketitle

\begin{abstract}
Recursive self-improvement (RSI) seeks to improve an agent using its own
task experience. Its practical value depends on whether performance gains
justify the full cost of self-improvement and task execution. We evaluate this
trade-off on continual task streams, scoring each task
before its feedback is used for subsequent updates.
We define the \emph{RSI Tax} as the token cost of the self-improvement
process and include it alongside task-execution costs in end-to-end evaluation. Across six
methods and two language models, a bounded-note baseline matches or
exceeds the observed accuracy of Meta-Harness, EvoSkill, and RRSI,
with mean total-token savings of \meanmetasaving{}--\meanevosaving{}\%.
We decompose the RSI Tax into verification, exploration, and writing
costs, and track subsequent use of
the associated state or artifact. For these three RSI methods,
verification accounts for 74.8--93.6\% of the aggregate RSI Tax. A substantial
share of the RSI Tax is also associated with states or artifacts
that are not used later in the evaluated streams. These results expose
a practical limitation of costly RSI procedures: improvements in the
deployed agent need not justify the expense of the improvement process.
They motivate RSI objectives that jointly optimize the RSI Tax and
expected task-execution cost while ensuring continued growth in agent
capabilities across the task stream.
\end{abstract}

\input{introduction}
\input{setup}
\input{results}
\input{discussion_conclusion}

\clearpage
\FloatBarrier
\begingroup
\small
\raggedright
\setlength{\bibsep}{3pt plus 1pt}

\input{main.bbl}
\endgroup

\clearpage
\appendix
\input{appendix}

\end{document}

%% file: introduction.tex
\section{Introduction}
\label{sec:intro}

Large language model (LLM) agents use tools to perform software
development and other professional tasks~\citep{claudecode,codex,terminalbench,apexagents}. During continued use, an agent may encounter related
tasks within the same repository or application.
Retaining useful procedures and lessons from failures could reduce
repeated exploration and improve later task performance. Recent sequential benchmarks examine this reuse
through connected workflows, related task variants, and skill
accumulation~\citep{cle,seaeval,continualskillbench}.

Recursive self-improvement (RSI) uses task experience to update the state
an agent carries into subsequent tasks. We focus on RSI with fixed model
weights and distinguish methods by the components they update. One category of RSI maintains an evolving set
of memories or skills for use in later tasks~\citep{reasoningbank,dynamiccheatsheet,evoskill}.
Another category modifies the agent's harness, including the instructions,
tool interfaces, and control code used for task execution~\citep{metaharness,rrsi,harnesspro}.

The self-improvement process has a cost of its own. Recent studies compare online
auxiliary modules and task-level search under resource constraints,
and relate capability gains to learning
cost~\citep{worth,rethinking,plasticity}. Validation-task selection and
the amortization of skill-search cost are also
studied~\citep{taskcoevolve,skillmoo}. On a task stream, each update incurs
a cost before its value for later tasks is known. Task
execution itself also varies in token demand across tasks, runs,
and models~\citep{agentspend}.

We call the token cost of the self-improvement process the \emph{RSI Tax}. We count the
tokens spent attempting an update even when the agent ultimately retains
its previous state. End-to-end cost is the sum of these update tokens
and the tokens used to execute the task stream. The two terms
are coupled: an update can shorten later executions, but can also add
context or induce longer trajectories. The relevant trade-off is between
the agent's capability gains over the task stream and this combined token cost.

\begin{figure}[!t]
\centering
\includegraphics[width=\linewidth]{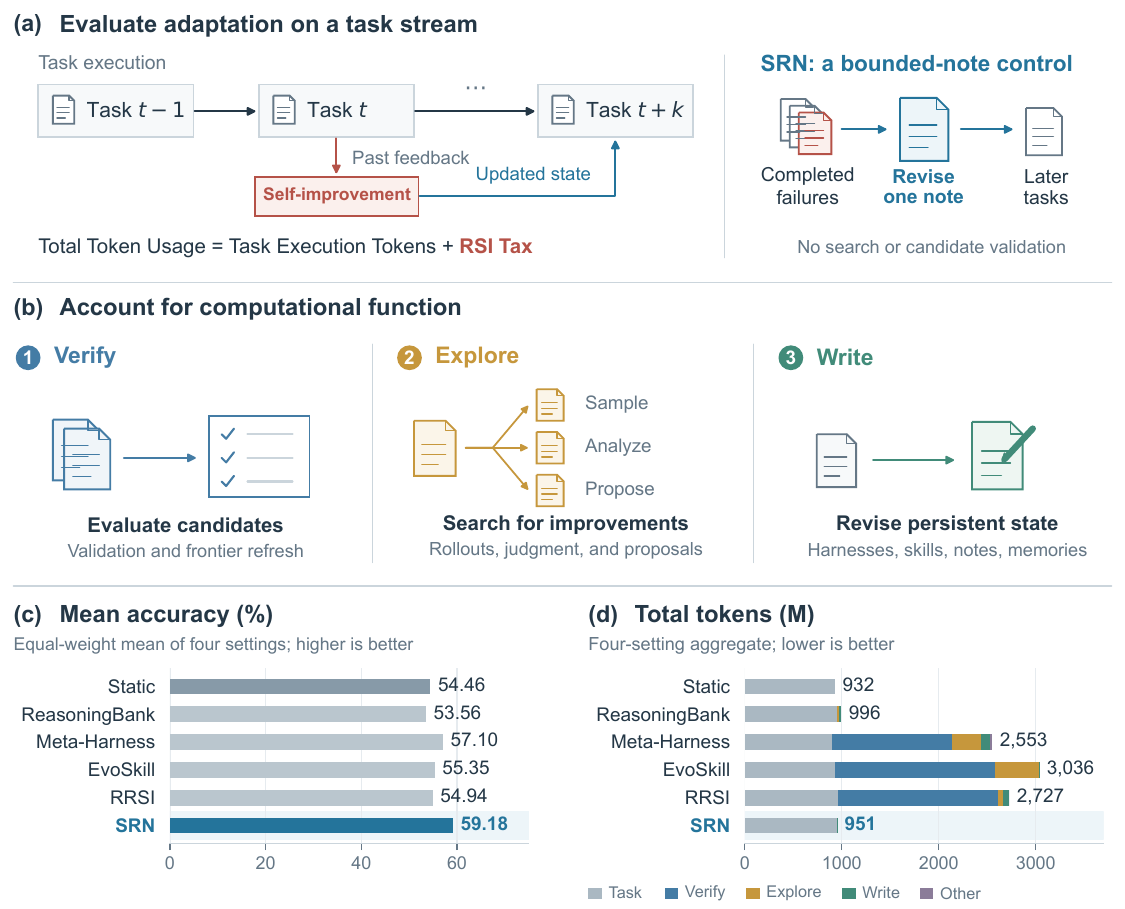}
\caption{\textbf{Streaming evaluation and the RSI Tax.}
(a) Updates use completed-task feedback; SRN maintains a bounded note.
(b) The RSI Tax is broken down into verification, exploration, and
writing costs. (c) Accuracy averaged equally over the four settings.
(d) Input-plus-output tokens summed over those settings, including task
execution, the three RSI Tax components, and the Other residual. These sums
describe token volume across the runs; within-setting comparisons are
reported in Section~\ref{sec:mainresults}.}
\label{fig:overview}
\end{figure}

We compare six methods on repository and terminal task streams using
two backbone models. Our simple baseline, Simple Reflection with Notes
(SRN), revises one bounded note from completed failures without
candidate search. Across the four model--environment settings, SRN
matches or exceeds the observed accuracy of Meta-Harness, EvoSkill,
and RRSI while using
\meanmetasaving{}--\meanevosaving{}\% fewer total tokens on average,
depending on the comparison method (Figure~\ref{fig:overview}).

To explain these differences, we decompose the RSI Tax into
\emph{verification}, \emph{exploration}, and \emph{writing}.
For these three RSI methods, verification accounts for 74.8--93.6\% of
the RSI Tax after aggregation over the four settings. We also track whether
the associated state or artifact is used by later tasks. Among the
evaluated RSI methods other than SRN, 23.7--63.6\% of the RSI Tax is
associated with states or artifacts that are not used later in the
observed stream.

Our results expose a gap between recursively improving an agent and making the
improvement process cost-effective.
Our decomposition shows that repeated verification can consume more
tokens than task execution itself, while substantial update costs are
associated with artifacts that are not used later. HarnessPro and RRSI
already account for execution cost in harness selection~\citep{harnesspro,rrsi}. Their
selection objectives do not, however, jointly optimize the cost of
obtaining updates and the expected cost of serving a continuing task
stream. Our findings therefore motivate policies that choose when to
update, how much to spend on proposals and validation, and which
artifacts to retain according to their expected contribution to future
task performance and token use. The goal is to sustain growth in agent
capabilities as tasks arrive while jointly optimizing the RSI Tax and
expected task-execution cost.

To summarize, our contributions are:
\begin{enumerate}[label=\arabic*.,topsep=4pt,itemsep=3pt]
\item \textbf{A streaming cost--performance evaluation.}
We compare six methods on common task streams and assess accuracy
against end-to-end token cost, including the RSI Tax.
\item \textbf{A simple adaptive baseline.}
SRN provides a counterexample to the need for costly update procedures to attain
the performance observed in these experiments.
\item \textbf{A decomposition of the RSI Tax.}
We separate verification, exploration, and writing costs, and link them
to subsequent use of the associated state or artifact.
\end{enumerate}

%% file: setup.tex
\section{Evaluation Setup}
\label{sec:problem}

\subsection{RSI Methods and Evaluation Protocols}
\label{sec:evaluation-context}
We study RSI that updates an agent's persistent state while keeping
its backbone model fixed. One category of RSI updates the memories or
skills available to the agent during task execution. ReasoningBank extracts strategies
from completed trajectories and retrieves them for later
tasks~\citep{reasoningbank}; Dynamic Cheatsheet maintains an evolving
note~\citep{dynamiccheatsheet}. EvoSkill searches over reusable skill
folders~\citep{evoskill}, while SkillOpt applies bounded edits to skill
documents~\citep{skillopt}.

Another category modifies the harness that controls prompting, tool use,
and task execution. GEPA
optimizes instructions using execution feedback~\citep{gepa}, and
Meta-Harness searches over controller code~\citep{metaharness}. RRSI
regularizes harness changes to discourage benchmark-specific
edits~\citep{rrsi}; HarnessPro considers execution cost as well as
performance when selecting a harness~\citep{harnesspro}.

Evaluation procedures differ in when feedback becomes available and
where performance is measured. ReasoningBank updates memory between
successive tasks, whereas EvoSkill and RRSI optimize skills or harnesses
on a fixed task collection before evaluating the selected state on
held-out tasks or other benchmarks~\citep{reasoningbank,evoskill,rrsi}.
EvoTest evaluates adaptation across repeated episodes of the same
game~\citep{evotest}, whereas TTHE adapts to an unlabeled current batch
before scoring performance on that batch~\citep{tthe}. ServeLearnBench separates serving and held-out
test tasks within each window and includes learning calls in its
resource accounting~\citep{servelearnbench}.

\subsection{Task Streams and Metrics}
\label{sec:stream}
We record a task's outcome before using its feedback to update the agent. Let $x_1,\ldots,x_T$ be the ordered tasks, $\theta$ the fixed
model parameters, and $s_t$ the state used for task $t$. Execution
produces a trajectory and a binary verification outcome:
\begin{equation}
\tau_t\sim\pi_\theta(\cdot\mid x_t,s_t),\qquad
 y_t=V(x_t,\tau_t)\in\{0,1\}.
\end{equation}
Updates use completed-task history
$\mathcal H_t=\{(x_i,\tau_i,y_i):i\leq t\}$, restricted by the state
boundaries and support windows. They may read public instructions,
visible trajectories, and scalar outcomes, but not future tasks,
reference solutions, or hidden test contents. Tokens spent re-executing
completed tasks are charged to the RSI Tax; these runs do not replace
the tasks' original outcomes.

Repository tasks come from seven selected 15-task chains in
SWE-Chain-Evo, totaling 105 positions~\citep{evoarena}.
Learned state persists within each chain and resets between chains;
each task starts from its prescribed repository snapshot. Terminal-Bench
2.1 supplies 89 heterogeneous tasks in a fixed order, with learned state
retained throughout~\citep{tb21}. Ordering these independent tasks allows
cross-task reuse but does not introduce environmental drift.

For a prefix of length $t$, we define
\begin{equation}
Q(t)=\sum_{i=1}^{t}y_i,\qquad A(t)=\frac{Q(t)}{t},\qquad
N(t)=N_{\rm task}(t)+N_{\rm RSI}(t).
\label{eq:streammetrics}
\end{equation}
$N(t)$ includes the tokens used for task execution and the RSI Tax
incurred through that position. We report final accuracy $A(T)$ and total tokens $N(T)$;
Section~\ref{sec:tax} specifies the token accounting. Sequential reuse
of experience overlaps with continual learning. Our objective is the
accuracy and cost of serving the arriving tasks, while concurrent work
on self-evolving agents evaluates retention across task
families~\citep{continuallearners}.

\subsection{Experimental Settings}
\label{sec:environments}
We compare Static, ReasoningBank, EvoSkill, Meta-Harness, RRSI, and SRN
(Section~\ref{sec:method}) using \dsmodel{}~\citep{deepseekv41} and
\lunamodel{}~\citep{gpt6luna}. This gives
$6\times2\times(105+89)=2{,}328$ scored method--task pairs. Within each
model--benchmark setting, the methods share the task order, initial
KIRA agent, and nominal execution limits.

SRN, Meta-Harness, and EvoSkill update every four completed tasks and
hold the selected state fixed within each window. Parallel tasks in a
window do not receive one another's feedback. Their support sets contain
the two most recent tasks on DeepSeek/SWE and four tasks in the other
settings. ReasoningBank retains per-task retrieval and memory updates,
without its separate memory-aware test-time scaling component.

The original EvoSkill procedure is adapted to changing support sets
by re-scoring its existing frontier. Our Meta-Harness selector requires
a strict accuracy improvement to change the deployment. We also retain
the methods' feedback choices: SRN selects failures using official
scalar outcomes, while ReasoningBank uses its internal trajectory judge.
Model parameters and implementation details are in
Appendix~\ref{app:protocol}.

\section{The RSI Tax}
\label{sec:tax}

\subsection{Cost Accounting}
For a recorded model call $c$, let $n(c)$ be its input-plus-output
usage. Task-execution calls $\mathcal C_{\rm task}$ and self-improvement calls
$\mathcal C_{\rm RSI}$ are disjoint. The absolute RSI Tax and its rate
are
\begin{equation}
\begin{aligned}
N_{\rm task}&=\sum_{c\in\mathcal C_{\rm task}}n(c),&
N_{\rm RSI}&=\sum_{c\in\mathcal C_{\rm RSI}}n(c),\\
\rho_{\rm RSI}&=\frac{N_{\rm RSI}}{N_{\rm task}},&
N_{\rm total}&=N_{\rm task}+N_{\rm RSI}.
\end{aligned}
\label{eq:accounting}
\end{equation}
The tax includes the tokens spent producing or evaluating an update,
regardless of whether that update is adopted. Reading learned state
during task execution belongs to $N_{\rm task}$. Embedding tokens are
included in the RSI Tax under Explore.
The RSI Tax accounts for $\rho_{\rm RSI}/(1+\rho_{\rm RSI})$ of total
token usage; at a Tax rate of 100\%, it equals task-execution cost. We compare token costs within each
model setting.

Relative to a reference method $b$, the net token premium is
\begin{equation}
\Delta N_{m,b}=N_{{\rm total},m}-N_{{\rm total},b}.
\label{eq:premium}
\end{equation}
We compare this premium with task accuracy. We next break down the RSI
Tax by operation and by subsequent use of the associated state or artifact.

\subsection{Decomposing RSI Tax}
\label{sec:functions}
We report verification, exploration, and writing separately, while
retaining the reported \emph{Other} residual (Table~\ref{tab:function-mapping}).

\paragraph{Verify.}
Verification evaluates an incumbent or candidate on completed support
tasks to inform selection. It includes baseline validation, candidate
smoke tests and further validation, and re-evaluation of an existing
skill frontier. Verification performed during the original task execution remains in
$N_{\rm task}$.

\paragraph{Explore.}
Exploration gathers or analyzes experience used to propose an update.
It includes extra task rollouts, proposal generation without artifact
edits, trajectory judgment, and ReasoningBank's embedding calls.

\paragraph{Write.}
Writing creates or revises persistent state, including harness code,
skills, notes, and memories. For Meta-Harness, trajectory analysis and
candidate search are charged to Explore, and candidate-code edits to
Write. EvoSkill likewise separates skill proposals from materialization.
SRN combines note analysis and revision in one operation, which is
charged entirely to Write. Each cost item is counted once.

\begin{table}[!htbp]
\caption{\textbf{RSI operations in the three cost categories.}
Token totals also retain a separate Other residual, reported in
Appendix~\ref{app:outcomes}.}
\label{tab:function-mapping}
\centering\small
\begin{tabularx}{\linewidth}{@{}l*{3}{>{\raggedright\arraybackslash}X}@{}}
\toprule
Method & Verify & Explore & Write \\
\midrule
ReasoningBank & --- & Trajectory judgment; embeddings & Memory extraction \\
\addlinespace
Meta-Harness & Incumbent and candidate validation & Trajectory analysis; candidate search & Candidate harness editing \\
\addlinespace
EvoSkill & Initial, new-skill, and frontier validation & Extra rollouts; skill proposals & Skill materialization \\
\addlinespace
RRSI & Incumbent and candidate validation & Trajectory analysis and summarization & Candidate harness generation and revision \\
\addlinespace
\notebaseline{} & --- & --- & Joint note analysis and revision \\
\bottomrule
\end{tabularx}
\end{table}

\subsection{Use in Subsequent Tasks}
\label{sec:downstream-impact}
We classify costs within the RSI Tax by whether the associated state or
artifact is used in a subsequent task before the next reset or the
end of the stream. Use includes running an updated harness, using a
skill during execution, or adding a revised note or retrieved memory
to the agent's task context. Verification and exploration costs
follow their recorded association with a state or artifact; attribution
is not limited to the final writing call. The labels record subsequent
use, not a measured effect on task behavior or accuracy.

Let $p(c)\in\{V,E,W,O\}$ identify the cost category of RSI call $c$
(verification, exploration, writing, or Other), and
$d(c)\in\{1,0\}$ the use status of its associated state or artifact. Then
\begin{equation}
N_{p,d}=\sum_{c\in\mathcal C_{\rm RSI}}n(c)
\,\mathbf{1}\{p(c)=p,\ d(c)=d\},\qquad
N_{\rm total}=N_{\rm task}+\sum_{p\in\{V,E,W,O\}}\sum_{d\in\{0,1\}}N_{p,d}.
\label{eq:function-impact}
\end{equation}
The share of the RSI Tax associated with unused state or artifacts is
\begin{equation}
r_{\rm no\text{-}use}=\frac{\sum_p N_{p,0}}{N_{\rm RSI}}.
\label{eq:noimpact}
\end{equation}
This is a share of tokens, not a fraction of updates or artifacts.
Section~\ref{sec:analysis} reports costs by category and use status
$N_{p,d}$, with category totals $N_p=\sum_d N_{p,d}$ and use-status
totals $N_d=\sum_p N_{p,d}$.
Appendix~\ref{app:impact-accounting} describes aggregation and precision,
and Appendix~\ref{app:case-studies} provides supplementary event diagnostics.

\section{Simple Reflection with Notes (SRN)}
\label{sec:method}
SRN is a bounded-note baseline based on cumulative-note
adaptation~\citep{dynamiccheatsheet}. At each four-task update, SRN uses
scalar verifier outcomes to select failed tasks for note revision; it
includes the current note in later task contexts. The note is limited to 2,048 tokens.

At an update boundary, let $m$ be the current note and
$\mathcal W_t$ the permitted recent window. The update is
\begin{equation}
\begin{aligned}
\mathcal F_t&=\{(x_i,\tau_i,y_i)\in\mathcal W_t:y_i=0\},\\
m^+&=\begin{cases}
m,&\mathcal F_t=\varnothing,\\
\operatorname{Limit}_{2048}\!\left(G_\theta(m,\operatorname{Excerpt}(\mathcal F_t))\right),
&\text{otherwise}.
\end{cases}
\end{aligned}
\label{eq:srn}
\end{equation}
The generator receives the previous note, public instructions, fixed
trajectory excerpts, and scalar outcomes. A valid nonempty revision is
adopted directly; an invalid response leaves the note unchanged.
Appendix~\ref{app:brief} gives the prompt and excerpt limits.

SRN performs no extra task sampling, candidate validation, or retrieval
search. Its total cost includes note construction and the complete
execution cost of solving tasks with the note. It provides a low-cost
adaptive comparison, although the experiment does not isolate the
effect of each design choice.

%% file: results.tex
\FloatBarrier
\section{Experiments}
\label{sec:results}
\suppressfloats[t]

We compare accuracy and total token cost, then break down the RSI Tax
by operation and subsequent use. Accounting details are given in
Appendix~\ref{app:provenance}.

\subsection{Accuracy and End-to-End Cost}
\label{sec:mainresults}
\begin{figure}[!t]
\centering
\begin{subfigure}[t]{0.485\linewidth}
\centering
\begin{tikzpicture}
\begin{axis}[costaxis,xmin=0,xmax=1250,ymin=28,ymax=54,xtick={0,400,800,1200},ytick={30,35,40,45,50}]
\costpoint{method0}{487.918178}{43.809524}{360}{44}{Static}
\costpoint{method3}{520.644331}{40.000000}{660}{37.5}{ReasoningBank}
\costpoint{method4}{976.608458}{45.714286}{855}{48.8}{Meta-Harness}
\costpoint{method5}{1166.514614}{40.952381}{1100}{38.5}{EvoSkill}
\costpoint{method2}{492.557891}{48.571429}{550}{51}{\textbf{SRN}}
\costpoint{method6}{1092.943785}{44.761905}{1120}{46.8}{RRSI}
\end{axis}
\end{tikzpicture}
\caption{DeepSeek V4.1 Flash / SWE-Chain-Evo}
\end{subfigure}
\hfill
\begin{subfigure}[t]{0.485\linewidth}
\centering
\begin{tikzpicture}
\begin{axis}[costaxis,xmin=0,xmax=1250,ymin=28,ymax=54,xtick={0,400,800,1200},ytick={30,35,40,45,50}]
\costpoint{method0}{179.685226}{38.095238}{105}{35}{Static}
\costpoint{method3}{191.212767}{37.142857}{430}{35}{ReasoningBank}
\costpoint{method4}{495.810084}{40.000000}{510}{43}{Meta-Harness}
\costpoint{method5}{612.387655}{40.000000}{900}{41.2}{EvoSkill}
\costpoint{method2}{157.071258}{40.952381}{250}{44.5}{\textbf{SRN}}
\costpoint{method6}{574.074999}{39.047619}{770}{36.5}{RRSI}
\end{axis}
\end{tikzpicture}
\caption{GPT-6 Luna / SWE-Chain-Evo}
\end{subfigure}
\par\medskip
\begin{subfigure}[t]{0.485\linewidth}
\centering
\begin{tikzpicture}
\begin{axis}[costaxis,xmin=0,xmax=1000,ymin=58,ymax=82,xtick={0,250,500,750,1000},ytick={60,65,70,75,80}]
\costpoint{method0}{185.549746}{69.662921}{90}{67.5}{Static}
\costpoint{method3}{198.869011}{69.662921}{410}{67.8}{ReasoningBank}
\costpoint{method4}{762.572645}{77.528090}{760}{80}{Meta-Harness}
\costpoint{method5}{898.722149}{71.910112}{860}{74}{EvoSkill}
\costpoint{method2}{214.652586}{78.651685}{290}{80}{\textbf{SRN}}
\costpoint{method6}{605.745029}{74.157303}{570}{76.7}{RRSI}
\end{axis}
\end{tikzpicture}
\caption{DeepSeek V4.1 Flash / Terminal-Bench 2.1}
\end{subfigure}
\hfill
\begin{subfigure}[t]{0.485\linewidth}
\centering
\begin{tikzpicture}
\begin{axis}[costaxis,xmin=0,xmax=1000,ymin=58,ymax=82,xtick={0,250,500,750,1000},ytick={60,65,70,75,80}]
\costpoint{method0}{78.864914}{66.292135}{90}{63.8}{Static}
\costpoint{method3}{85.461031}{67.415730}{190}{59.8}{ReasoningBank}
\costpoint{method4}{317.993991}{65.168539}{700}{64.5}{Meta-Harness}
\costpoint{method5}{358.852662}{68.539326}{560}{70}{EvoSkill}
\costpoint{method2}{87.077040}{68.539326}{240}{68.2}{\textbf{SRN}}
\costpoint{method6}{454.299580}{61.797753}{600}{59.7}{RRSI}
\end{axis}
\end{tikzpicture}
\caption{GPT-6 Luna / Terminal-Bench 2.1}
\end{subfigure}
\caption{\textbf{Accuracy and total token cost on the four task streams.}
Each point summarizes one method; total cost combines task-execution
cost and the RSI Tax. Higher accuracy and lower cost are preferable. Axes are shared within
each benchmark, and lines connect labels to points.}
\label{fig:tradeoff}
\end{figure}
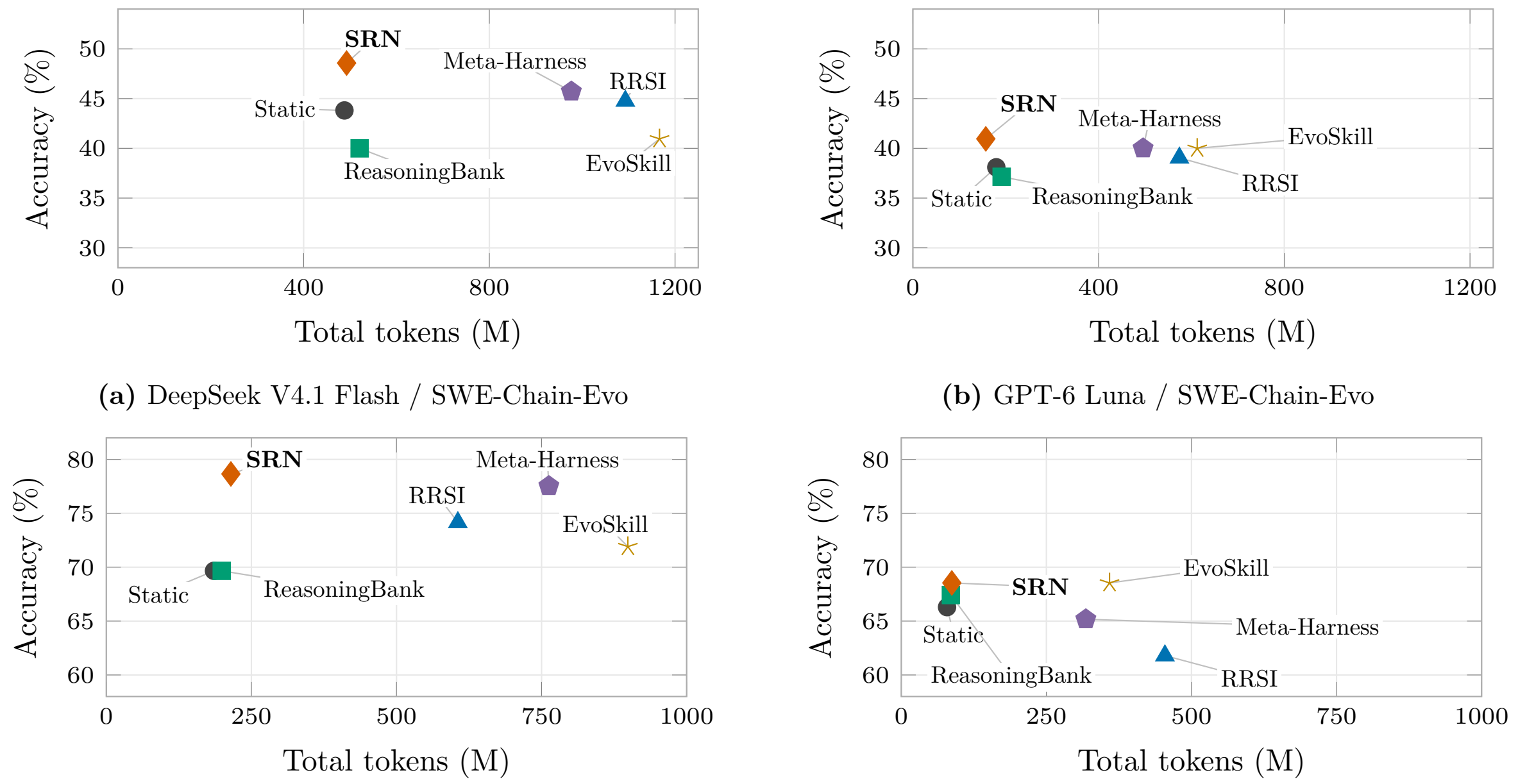

Table~\ref{tab:main-results} and Figure~\ref{fig:tradeoff} compare the six
methods. SRN has the highest observed accuracy in three settings and
ties EvoSkill in the fourth, while using fewer total tokens than
Meta-Harness, EvoSkill, and RRSI in every setting. ReasoningBank incurs a
smaller RSI Tax than these three methods but does not consistently
improve accuracy over Static.

\begin{table}[H]
\centering\small\setlength{\tabcolsep}{3pt}
\caption{\textbf{Accuracy and total token cost.}
Accuracy is in percent; Total is in millions of input-plus-output
tokens and includes both task-execution cost and the RSI Tax. DeepSeek denotes
DeepSeek V4.1 Flash. Bold marks the highest accuracy and lowest total cost in each setting, including ties. Detailed subtotals and accounting conventions
are in Appendices~\ref{app:totals} and~\ref{app:provenance}.}
\label{tab:main-results}
\begin{tabular*}{\linewidth}{@{\extracolsep{\fill}}lrrrrrrrr@{}}
\toprule
& \multicolumn{4}{c}{SWE-Chain-Evo (105 positions)}
& \multicolumn{4}{c}{Terminal-Bench 2.1 (89 tasks)} \\
\cmidrule(lr){2-5}\cmidrule(l){6-9}
& \multicolumn{2}{c}{DeepSeek} & \multicolumn{2}{c}{GPT-6 Luna}
& \multicolumn{2}{c}{DeepSeek} & \multicolumn{2}{c}{GPT-6 Luna} \\
\cmidrule(lr){2-3}\cmidrule(lr){4-5}\cmidrule(lr){6-7}\cmidrule(l){8-9}
Method & Acc. & Total & Acc. & Total & Acc. & Total & Acc. & Total \\
\midrule
Static & 43.8 & \textbf{487.92} & 38.1 & 179.69 & 69.7 & \textbf{185.55} & 66.3 & \textbf{78.86} \\
ReasoningBank & 40.0 & 520.64 & 37.1 & 191.21 & 69.7 & 198.87 & 67.4 & 85.46 \\
Meta-Harness & 45.7 & 976.61 & 40.0 & 495.81 & 77.5 & 762.57 & 65.2 & 317.99 \\
EvoSkill & 41.0 & 1,166.51 & 40.0 & 612.39 & 71.9 & 898.72 & \textbf{68.5} & 358.85 \\
RRSI & 44.8 & 1,092.94 & 39.0 & 574.07 & 74.2 & 605.75 & 61.8 & 454.30 \\
\midrule
SRN & \textbf{48.6} & 492.56 & \textbf{41.0} & \textbf{157.07} & \textbf{78.7} & 214.65 & \textbf{68.5} & 87.08 \\
\bottomrule
\end{tabular*}
\end{table}

Relative to Static, SRN gains \meanaccgain{} percentage points in
mean accuracy with a \meancostincrease{}\% mean increase in total
tokens. The mean token reductions relative to Meta-Harness, EvoSkill,
and RRSI are \meanmetasaving{}\%, \meanevosaving{}\%, and
\meanrrsisaving{}\%, respectively. These summaries weight the four settings equally: accuracy gains
are averaged differences, while token changes are computed as
within-setting ratios before averaging. SRN's update calls account for less than 0.4\% of its total tokens;
task execution, including use of the note, accounts for the rest.

The relation between accuracy and total token cost varies across settings.
On the DeepSeek terminal stream, Static uses the fewest tokens
(185.55M), while SRN uses 15.7\% more and gains 9.0 percentage points
in observed accuracy. On the GPT-6 Luna repository stream, SRN uses
156.67M task-execution tokens compared with Static's 179.69M, a
difference that more than offsets its 0.40M update cost. Its
total is 12.6\% lower than Static's and its accuracy is 2.9 points
higher.

\paragraph{Execution cost and total cost.}
Omitting the RSI Tax can reverse the cost comparison. On the DeepSeek
terminal stream, Meta-Harness uses fewer task-execution tokens than
SRN (184.13M versus 214.37M). Its 578.45M RSI Tax, however,
raises the total to 762.57M, compared with SRN's 214.65M, while its
observed accuracy is slightly lower (77.5\% versus 78.7\%). These task
totals cover all deployed states used during the stream, rather than
a separate evaluation of the final checkpoint.

\subsection{Cost Breakdown and Subsequent Use}
\label{sec:analysis}
The RSI Tax exceeds task-execution cost for Meta-Harness, EvoSkill,
and RRSI in all four settings (Table~\ref{tab:tax}). For \notebaseline{},
the tax rate remains below 0.4\%.

\begin{table}[!htbp]

\caption{\textbf{RSI Tax rate}, $100N_{\rm RSI}/N_{\rm task}$ (\%).
A value of 100 means that the RSI Tax equals task-execution cost. Static has rate zero. RRSI rates are the reported summary
values; the other rates are computed from unrounded token totals.
DeepSeek denotes DeepSeek V4.1 Flash. See Appendix~\ref{app:impact-accounting}
for the aggregation details.}
\label{tab:tax}

\centering\small
\begin{tabular*}{\linewidth}{@{\extracolsep{\fill}}lrrrr@{}}
\toprule
& \multicolumn{2}{c}{SWE-Chain-Evo} & \multicolumn{2}{c}{Terminal-Bench 2.1} \\
\cmidrule(lr){2-3}\cmidrule(l){4-5}
Method & DeepSeek & GPT-6 Luna & DeepSeek & GPT-6 Luna \\
\midrule
ReasoningBank & 4.78 & 5.96 & 4.53 & 7.39 \\
Meta-Harness & 109.06 & 177.35 & 314.16 & 329.15 \\
EvoSkill & 144.35 & 213.21 & 393.10 & 358.57 \\
RRSI & 125.79 & 213.02 & 223.23 & 346.68 \\
\notebaseline{} & 0.04 & 0.25 & 0.13 & 0.36 \\
\bottomrule
\end{tabular*}
\end{table}

\begin{figure}[!t]
\centering
\includegraphics[width=\linewidth]{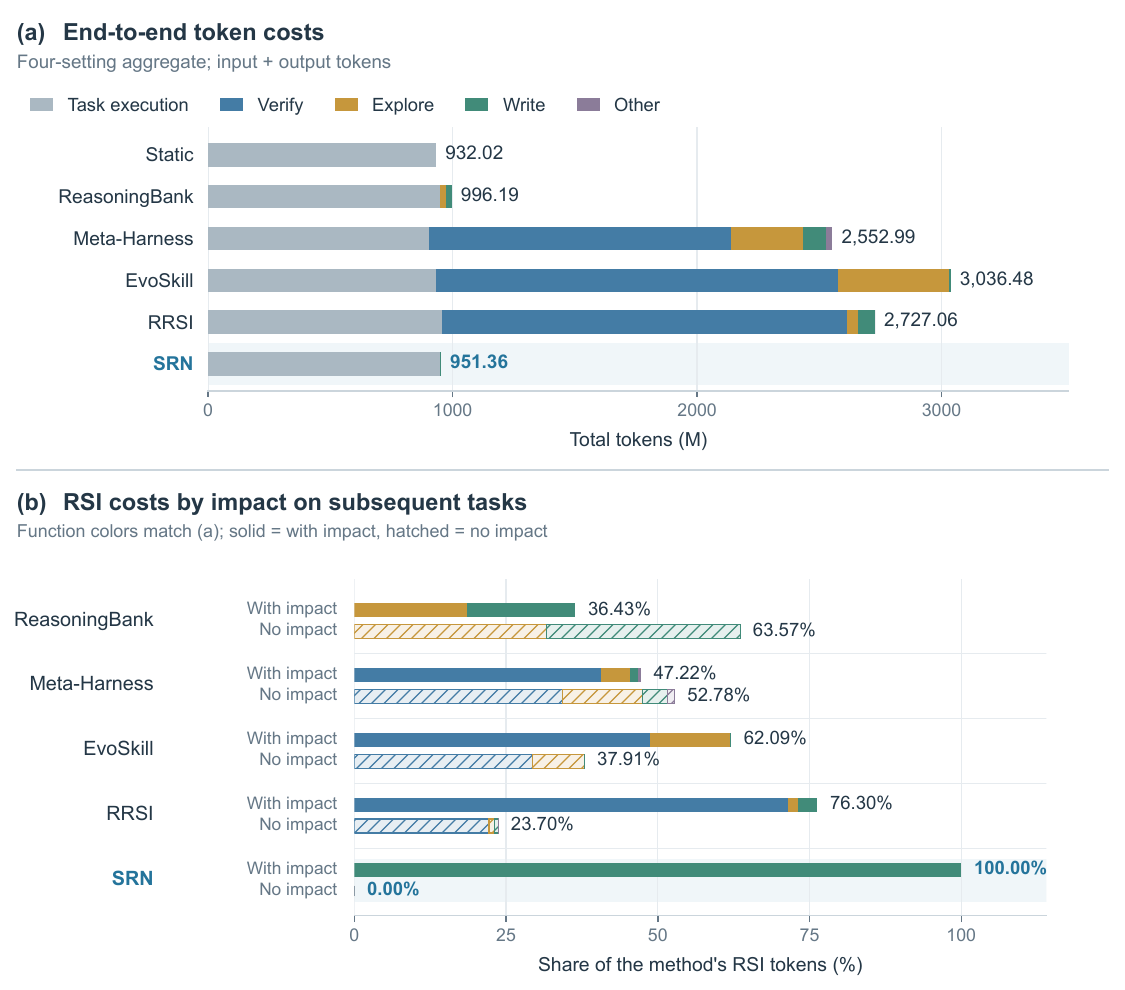}
\caption{\textbf{Token cost breakdown and subsequent use.}
(a) Input-plus-output tokens summed over the four settings, split into
task execution, verification, exploration, writing, and Other.
(b) RSI Tax associated with states or artifacts used in later
tasks (solid) or not used within the observation period (hatched).
Both rows are normalized by the method's total RSI Tax.
Static incurs no RSI Tax. These pooled token volumes are descriptive;
the mean cost comparisons in Section~\ref{sec:mainresults} use
within-setting ratios.}
\label{fig:cost-decomposition}
\end{figure}

Figure~\ref{fig:cost-decomposition} breaks down total token cost and
links the RSI Tax to subsequent use of the associated state or artifact.

\paragraph{Sources of the RSI Tax.}
Verification accounts for 74.8\% of Meta-Harness's RSI Tax,
78.1\% of EvoSkill's, and 93.6\% of RRSI's after aggregation over the
four settings. EvoSkill also spends 455.29M tokens on additional
rollouts and proposals, compared with 4.59M on skill materialization.
ReasoningBank divides its cost almost equally between exploration
(judgment and embeddings) and memory writing. SRN spends 1.21M tokens
on its joint note-analysis and revision operation. Full component
totals are given in Appendix~\ref{app:outcomes}.

Verification repeatedly runs the agent on support tasks, often evaluating
several candidate or retained programs per update. On the DeepSeek terminal stream, EvoSkill spends 497.45M tokens
refreshing its old frontier, compared with 182.26M on the 89 arriving
tasks. Re-evaluating existing programs alone therefore costs more than
serving the task stream, before accounting for extra rollouts or
new-skill validation. Recorded input accounts for approximately 98.1\%
of the RSI Tax in both Meta-Harness and EvoSkill.

\paragraph{Costs associated with subsequent use.}
Under the attribution in Section~\ref{sec:downstream-impact},
52.78\% of Meta-Harness's RSI Tax, 37.91\% of EvoSkill's,
63.57\% of ReasoningBank's, and 23.70\% of RRSI's are associated with
states or artifacts that are not used in later tasks. All reported SRN
update cost is associated with notes supplied to subsequent executions.
These percentages describe token expenditure, not the frequency of
successful updates or an effect on accuracy. Appendix~\ref{app:phases}
gives the component totals and Appendix~\ref{app:case-studies} examines
individual update traces.

The same lack of subsequent use can involve different operations.
Meta-Harness assigns 564.85M verification tokens, 217.60M exploration
tokens, and 69.22M writing tokens to candidates not used later.
For ReasoningBank, the corresponding costs are 15.51M for judgment
and embeddings and 15.63M for memory extraction. The breakdown identifies
different sources of RSI Tax in these two methods; their overall shares
of costs without subsequent use do not reveal this composition.

%% file: discussion_conclusion.tex
\FloatBarrier
\section{Discussion}
\label{sec:discussion}

The results show that the RSI Tax can exceed the cost of executing the
task stream. We discuss how this changes the design of update
policies, where costs might be reduced, and what the current evidence
can establish.

\paragraph{Choosing an update policy.}
An RSI update policy should pursue continued gains in agent capabilities
as tasks arrive while controlling the RSI Tax. Its efficiency objective should include the immediate
update cost, expected task-execution cost, and the cost of subsequent
RSI updates over the remaining tasks. Such a policy could retain the current state,
update memory or skills, or generate and evaluate harness changes.
Repeated reuse may justify a larger initial investment~\citep{skillmoo}.
Our experiments compare fixed procedures and realized costs; they do
not evaluate a policy that forecasts these costs and selects updates.

\paragraph{Reducing the RSI Tax.}
When verification dominates the RSI Tax, reducing the cost of candidate
evaluation becomes a priority.
Support-task selection and frontier maintenance are therefore
natural targets for reducing update cost. Adaptive validation-task
selection already provides one approach~\citep{taskcoevolve}.
The subsequent-use analysis also motivates more selective construction
of skills and memories. Whether these changes reduce total tokens
while preserving task performance requires controlled evaluation.
Reducing update cost complements the execution-cost objectives used in
cost-aware harness selection~\citep{rrsi,harnesspro}.

\paragraph{Limitations.}
The evaluation covers two backbone models, seven repository chains,
and one ordering of a terminal benchmark. Other models, workloads,
orderings, and repeated runs are needed to assess the stability of the
observed comparisons. The cost attribution is descriptive: use in a
later task does not establish an accuracy benefit, and the absence of
such use does not establish that the associated calls can be removed
without changing later updates or task outcomes. Controlled interventions
are needed to measure the effects of reducing each cost component.

\section{Conclusion}
\label{sec:conclusion}
We evaluate the end-to-end token cost and task performance of mainstream
RSI methods on task streams, using SRN as a simple adaptive baseline.
We introduce the RSI Tax to quantify the token cost of the
self-improvement process during deployment.
On the studied streams, SRN
shows that competitive task accuracy need not require a large RSI Tax.
Decomposing the RSI Tax into verification, exploration, and writing costs
shows that repeated verification can consume more tokens than executing
the task stream itself.
A substantial share of the RSI Tax is also associated with states or
artifacts that are not used in subsequent tasks. These findings motivate
RSI agents that jointly optimize the RSI Tax incurred by trial-and-error
self-improvement and expected task-execution cost while ensuring
continued growth in agent capabilities across the task stream.

%% file: appendix.tex
\section{Experimental Protocol}
\label{app:protocol}

\subsection{Streams, State Boundaries, and Model Parameters}
The reported SWE-Chain-Evo subset uses Click, CoreDNS, NATS, Fiber,
Loki, Istio, and Prometheus, each with 15 ordered task positions.
These 105 positions are a selected subset of EvoArena, whose full
SWE-Chain-Evo benchmark contains 50 chains and 493 chain-step
instances~\citep{evoarena}. Notes, memories,
skills, and harness state reset at chain boundaries. Each task starts
from its benchmark-prescribed repository snapshot; earlier agent patches
are not transplanted into later tasks. The current repository and public
history remain available to all methods, including Static. The terminal
evaluation uses all 89 Terminal-Bench 2.1 tasks in one fixed order with
task-specific environments and persistent learned state across the stream.

The recorded router identifiers are \texttt{deepseek/deepseek-v4.1-flash}
and \texttt{openai/gpt-6-luna}. Official model documentation is provided
by \citet{deepseekv41,gpt6luna}. The recorded harness configuration specifies
high reasoning effort for both models, a nominal
262,144-token context limit, a nominal 65,536-token per-call output
limit, and at most 100 agent turns, without a cumulative per-task
token cap. These are harness-imposed limits, not the maximum capacities
advertised by the providers. DeepSeek uses temperature 1 and top-$p$ 0.95;
the reported GPT-6 Luna configuration omits both sampling parameters.
Model-specific tokenization
means that nominal limits do not imply identical effective capacity.
SWE solving and verification limits are 7,200 and 600 seconds;
terminal tasks use their official task-specific limits.

\subsection{Periodic Updates and Support Sets}
\notebaseline{}, Meta-Harness, and EvoSkill have an update opportunity
after every four completed tasks, with no final partial-window update.
There are three opportunities per repository chain and 22 per terminal
stream. Periodic methods freeze the selected state within each four-task
window; parallel execution does not expose another task's feedback inside
that window. ReasoningBank retains its per-task memory workflow.

The DeepSeek V4.1 Flash repository experiments use the most recent two
tasks as update support; the other three settings use the most recent
four. Thus, the four-task \emph{update interval} is shared, while support
size differs between those settings. Comparisons are made within each
setting before macro averaging; support size is held fixed within a
setting. Re-evaluating support
tasks supplies scalar feedback for future updates, not replacements for
their earlier formal outcomes.

\subsection{Method Configurations}
Static retains the initial KIRA execution policy and no learned
cross-task state. The reported Meta-Harness configuration permits one
native search iteration per update, with smoke evaluation and support-set
validation. Its deployment adapter requires a strict improvement in
mean validation accuracy and retains the incumbent on a tie. Subsequent
tasks run under the selected harness. This rule is narrower
than the population-based search described by the original
method~\citep{metaharness}; the saturated-selector case concerns this
evaluated deployment rule.

RRSI's update procedure includes an analyst and trajectory digesters,
a proposer, a critic, and validation.
The analyst and digesters examine past trajectories and summarize
failure modes and reusable behavior; the proposer constructs candidate
harness edits.
The critic separately screens candidate changes before validation~\citep{rrsi}.

The reported EvoSkill configuration uses the upstream proposal, skill
materialization, and program-selection operators with one native
iteration, a frontier size of three, and one sampling category containing
two samples. Existing frontier
members are re-scored on the current support set as the window moves.
The original algorithm uses a fixed validation set and evaluates new
candidates~\citep{evoskill}. The rolling refresh is part of our streaming
implementation and its full recorded cost is charged. The original
proposer may inspect ground-truth answers; our information boundary permits
only public instructions, visible traces, and scalar feedback from
completed tasks. The selected skill catalog is made available to the
agent through the existing skill-discovery interface.
These runs use the original EvoSkill procedure; EvoSkill v2 instead
adopts each epoch's skill revision without validation-based
rejection~\citep{evoskillv2}.

ReasoningBank preserves trajectory judgment, memory extraction, and
retrieval of the highest-ranked episode bundle. A bundle can contain
multiple memory items; retrieving it adds those items to the agent's task
context. ReasoningBank's trajectory judge is distinct from the task
verifier used to calculate accuracy. Embedding uses \texttt{gemini-embedding-001};
embedding tokens are included in the RSI Tax. This evaluates the
memory workflow without adding a separate inference-scaling treatment.

\subsection{Note Baseline and Prompt}
\label{app:brief}
\notebaseline{} maintains a note of at most 2,048 tokens and uses the same
backbone for note updates and task execution. The generator receives
the previous note, public instructions, scalar outcomes, and excerpts
from failed trajectories. Each excerpt contains the first 20,000 and
last 10,000 characters, or the full trajectory when shorter. A valid,
nonempty response replaces the note; an invalid response leaves it
unchanged. Oversized notes are truncated by the configured tokenizer.
Each update uses one logical generation, and all retry calls remain
in the token count.

The terminal note prompt is reproduced below; the repository prompt
uses the same constraints in the repository-chain setting.
\begin{quote}\small
\begin{Verbatim}[fontsize=\footnotesize]
Maintain a short experience note for a terminal agent learning across
a chronological sequence of independent tasks. Use only the supplied
previous note, public task instructions, visible execution traces, and
scalar verifier rewards.

Return only the replacement note in plain Markdown. Keep the final
note within 2048 tokens. Keep useful earlier lessons, remove outdated
or unsupported claims, and add reusable lessons supported by the
current evidence. Prefer concrete environment inspection, tool-use,
debugging, implementation, and verification advice. Distinguish
observed facts from hypotheses. Do not infer hidden tests or reference
solutions, invent command results, or include instructions to access
future tasks. Do not solve the next task, request tools, or repeat a
full task-specific patch. Experience is guidance; future tasks may
change the relevant assumptions.

Trace excerpts use a fixed head/tail view and may omit intermediate
steps. A scalar reward of zero does not establish a specific root
cause. Avoid drawing strong causal lessons without supporting visible
evidence. An empty useful update may retain the previous note. Output
a note, without surrounding explanation.
\end{Verbatim}
\end{quote}

\section{Accounting and Interpretation}
\label{app:accounting}
\subsection{Token and Outcome Accounting}
\label{app:provenance}
The task-execution total includes context management and reading retained state.
The RSI Tax includes tokens spent on memory construction, embeddings,
proposal sessions, extra sampling, and candidate and incumbent validation. Request
identities prevent double-counting across logs. Input and output tokens
are included; cached input remains in the count, and reasoning tokens
already included in provider output are not added a second time.
The per-setting stage sums reconcile with the corresponding RSI totals
in Table~\ref{tab:full-costs}, subject to display rounding.
These totals measure token use rather than monetary cost, which also
depends on the pricing of input, output, and cached tokens~\citep{tokenreduction}.

Accuracy summarizes task outcomes under the streaming protocol in
Section~\ref{sec:stream}. Tokens spent on support-set validation are
charged to the RSI Tax; validation scores do not replace the earlier task outcomes.

\subsection{Cross-Setting Summaries}
Let $A_{m,g}$ denote fractional accuracy and $N_{m,g}$ total tokens for
method $m$ in setting $g$. For comparison method $b$, we report
\begin{equation}
\Delta A_b=\frac{100}{4}\sum_{g=1}^{4}(A_{\mathrm{\notebaseline{}},g}-A_{b,g}),
\qquad
S_b=\frac{100}{4}\sum_{g=1}^{4}\left(1-\frac{N_{\mathrm{\notebaseline{}},g}}{N_{b,g}}\right).
\end{equation}
The unrounded records give $\Delta A_{\rm Static}=4.714$ percentage points,
$S_{\rm Static}=-3.616\%$, $S_{\rm Meta\text{-}Harness}=65.588\%$,
$S_{\rm EvoSkill}=70.994\%$, and $S_{\rm RRSI}=68.242\%$.
The mean accuracy difference relative to RRSI is 4.238 percentage points. Each model--benchmark setting receives equal weight. Token ratios are
formed within settings before averaging.

\subsection{Cost Categories and Subsequent Use}
\label{app:impact-accounting}
We report verification, exploration, and writing costs alongside the
Other residual from the source records, and separately track whether
the associated state or artifact is used in later tasks. Figure~\ref{fig:cost-decomposition}(b)
reports the resulting breakdown, with every segment normalized by
the method's complete RSI Tax.

The accounting is a sum over the four model--environment settings,
in millions of input-plus-output tokens. Summing the available per-setting
stage totals reproduces the corresponding aggregate RSI Tax, up to
display rounding.
This aggregate measures token volume, not a common
physical-compute unit across different model tokenizers. Comparative
ratios in the main results are computed within settings before averaging.

Cells, totals, and percentages are independently rounded, generally
to two decimal places; RRSI's Other cost is reported as
0.009M (less than 0.01\% of the RSI Tax). Independently rounded cells
need not sum exactly to the displayed totals. In particular,
Meta-Harness's use-status rows sum to 297.00M Explore and 91.73M Write,
while the separately reported category totals are 296.99M and 91.74M.
RRSI's 0.009M Other cost belongs to the row without subsequent use.
RRSI's aggregate RSI Tax is 1,770.50M tokens: 1,350.81M with subsequent
use and 419.69M without subsequent use.
For methods with separately reported full-precision total and RSI
counts, task tokens are computed as their difference. RRSI's aggregate
task-execution cost is approximately 956.56M tokens, computed as
2,727.06M total tokens minus the rounded 1,770.50M RSI total and subject
to that rounding precision.

The operation-to-category mapping is given in
Table~\ref{tab:function-mapping}.
Meta-Harness's proposal sessions total 414.969432M tokens, comprising
296.99M Explore, 91.74M Write, and 26.24M Other; these amounts sum
to 414.97M at the reported precision. The per-setting stage tables below
report proposal sessions as a combined category, while
Table~\ref{tab:impact-costs} gives their aggregate cost breakdown.
\notebaseline{}'s joint note-analysis and revision calls belong to Write.

\subsection{Linking Costs to Subsequent Execution}
We attribute each cost item to the state or artifact it concerns,
including items incurred during verification or exploration. For a
harness, subsequent use means execution under the associated controller;
for skills, notes, and memories, it means their use during task execution
or inclusion in the agent's context. Each cost item contributes once to the table; shared
verification or analysis costs follow the designated candidate. Attribution
follows the recorded association and subsequent use before reset or
the end of the stream. A round-level comparison of the deployed state
alone does not give the same classification, because a round may inspect
multiple artifacts. Use is recorded separately from changes in accuracy.

\section{Detailed Token Results}
\label{app:phases}
\subsection{Task Tokens and RSI Tax}
\label{app:totals}
Table~\ref{tab:full-costs} gives the task-execution costs and RSI Tax underlying
Table~\ref{tab:main-results}. Static incurs no RSI Tax, shown as an em
dash, and is excluded from the RSI-column minimum. RRSI's task-execution
cost and RSI Tax are reported at the aggregate level in
Figure~\ref{fig:cost-decomposition}; the Tax components appear in
Table~\ref{tab:impact-costs}.

\begin{table}[!htbp]
\centering\small\setlength{\tabcolsep}{3pt}
\caption{\textbf{Task-execution cost and RSI Tax.}
Accuracy is in percent; token costs are in millions. Static incurs no
RSI Tax and is excluded from the RSI-column minimum. Bold
marks the best reported value within each setting, including ties.
Display rounding can make Task plus RSI
differ slightly from Total.}
\label{tab:full-costs}
\begin{tabular*}{\linewidth}{@{\extracolsep{\fill}}lrrrrrrrr@{}}
\toprule
& \multicolumn{4}{c}{DeepSeek V4.1 Flash} & \multicolumn{4}{c}{GPT-6 Luna} \\
\cmidrule(lr){2-5}\cmidrule(l){6-9}
Method & Acc. & Task & RSI & Total & Acc. & Task & RSI & Total \\
\midrule
\multicolumn{9}{@{}l}{\textit{SWE-Chain-Evo (105 positions)}} \\
\addlinespace[2pt]
Static & 43.8 & 487.92 & \textemdash & \textbf{487.92} & 38.1 & 179.69 & \textemdash & 179.69 \\
ReasoningBank & 40.0 & 496.91 & 23.74 & 520.64 & 37.1 & 180.46 & 10.75 & 191.21 \\
Meta-Harness & 45.7 & \textbf{467.13} & 509.47 & 976.61 & 40.0 & 178.77 & 317.04 & 495.81 \\
EvoSkill & 41.0 & 477.40 & 689.12 & 1,166.51 & 40.0 & 195.52 & 416.87 & 612.39 \\
RRSI & 44.8 & 484.05 & 608.89 & 1,092.94
     & 39.0 & 183.40 & 390.68 & 574.07 \\
\addlinespace[2pt]
\notebaseline{} & \textbf{48.6} & 492.34 & \textbf{0.22} & 492.56 & \textbf{41.0} & \textbf{156.67} & \textbf{0.40} & \textbf{157.07} \\
\midrule
\multicolumn{9}{@{}l}{\textit{Terminal-Bench 2.1 (89 tasks)}} \\
\addlinespace[2pt]
Static & 69.7 & 185.55 & \textemdash & \textbf{185.55} & 66.3 & 78.86 & \textemdash & \textbf{78.86} \\
ReasoningBank & 69.7 & 190.26 & 8.61 & 198.87 & 67.4 & 79.58 & 5.88 & 85.46 \\
Meta-Harness & 77.5 & 184.13 & 578.45 & 762.57 & 65.2 & \textbf{74.10} & 243.89 & 317.99 \\
EvoSkill & 71.9 & \textbf{182.26} & 716.46 & 898.72 & \textbf{68.5} & 78.25 & 280.60 & 358.85 \\
RRSI & 74.2 & 187.40 & 418.34 & 605.75
     & 61.8 & 101.71 & 352.59 & 454.30 \\
\addlinespace[2pt]
\notebaseline{} & \textbf{78.7} & 214.37 & \textbf{0.28} & 214.65 & \textbf{68.5} & 86.77 & \textbf{0.31} & 87.08 \\
\bottomrule
\end{tabular*}

\end{table}

\subsection{Aggregate Costs and Subsequent Use}
\label{app:outcomes}
Table~\ref{tab:impact-costs}(a) reports the three RSI Tax components and
the Other residual; panel (b) groups these costs by subsequent use. Percentages use each method's
complete RSI Tax as the denominator. Static incurs no RSI Tax.

\input{rsi-impact-table}

\subsection{RSI Tax by Implementation Stage}

The following tables report token use by implementation stage for each
model--benchmark setting. Values include input and output usage. The
aggregate cost attribution is described in
Appendix~\ref{app:impact-accounting}; the aggregate visualization appears in
Figure~\ref{fig:cost-decomposition}. These stage totals do not themselves
determine the subsequent-use status of individual cost items.

\begin{table}[!htbp]\centering\small
\caption{Meta-Harness: RSI Tax by stage (millions of tokens). DeepSeek denotes DeepSeek V4.1 Flash.}
\begin{tabular*}{\linewidth}{@{\extracolsep{\fill}}lrrrr@{}}
\toprule
 & \multicolumn{2}{c}{SWE-Chain-Evo} & \multicolumn{2}{c}{Terminal-Bench 2.1} \\
\cmidrule(lr){2-3}\cmidrule(l){4-5}
Phase & DeepSeek & GPT-6 Luna & DeepSeek & GPT-6 Luna \\
\midrule
Proposal session & 109.997932 & 62.540720 & 155.750468 & 86.680312 \\
Incumbent validation & 193.403359 & 139.638809 & 211.104585 & 70.721386 \\
Candidate smoke & 95.848101 & 27.941626 & 38.117662 & 11.285732 \\
Further candidate validation & 110.224752 & 86.921894 & 173.474330 & 75.207386 \\
\midrule
Total & 509.474144 & 317.043049 & 578.447045 & 243.894816 \\
\bottomrule\end{tabular*}

\end{table}
\begin{table}[!htbp]\centering\small
\caption{EvoSkill: RSI Tax by stage (millions of tokens). DeepSeek denotes DeepSeek V4.1 Flash.}
\begin{tabular*}{\linewidth}{@{\extracolsep{\fill}}lrrrr@{}}
\toprule
 & \multicolumn{2}{c}{SWE-Chain-Evo} & \multicolumn{2}{c}{Terminal-Bench 2.1} \\
\cmidrule(lr){2-3}\cmidrule(l){4-5}
Phase & DeepSeek & GPT-6 Luna & DeepSeek & GPT-6 Luna \\
\midrule
Initial baseline validation & 55.390798 & 38.509148 & 7.343709 & 1.632162 \\
Extra rollouts & 200.396353 & 71.161754 & 112.474010 & 35.706044 \\
Skill proposal & 20.072240 & 6.416161 & 6.122136 & 2.946065 \\
Skill materialization & 1.361587 & 1.229830 & 1.092549 & 0.901379 \\
New-skill validation & 141.891026 & 120.816937 & 91.976435 & 43.580524 \\
Old-frontier refresh & 270.004290 & 178.735837 & 497.454675 & 195.831597 \\
\midrule
Total & 689.116294 & 416.869667 & 716.463514 & 280.597771 \\
\bottomrule\end{tabular*}

\end{table}
\begin{table}[!htbp]\centering\small
\caption{ReasoningBank: RSI Tax by stage (millions of tokens). DeepSeek denotes DeepSeek V4.1 Flash.}
\begin{tabular*}{\linewidth}{@{\extracolsep{\fill}}lrrrr@{}}
\toprule
 & \multicolumn{2}{c}{SWE-Chain-Evo} & \multicolumn{2}{c}{Terminal-Bench 2.1} \\
\cmidrule(lr){2-3}\cmidrule(l){4-5}
Phase & DeepSeek & GPT-6 Luna & DeepSeek & GPT-6 Luna \\
\midrule
Trajectory judgment & 11.578243 & 5.354226 & 4.659062 & 2.892283 \\
Memory extraction & 12.142389 & 5.378519 & 3.904759 & 2.943139 \\
Embeddings (combined) & 0.015485 & 0.015418 & 0.048303 & 0.048426 \\
\midrule
Total & 23.736117 & 10.748163 & 8.612124 & 5.883848 \\
\bottomrule\end{tabular*}

\end{table}

\paragraph{Notes and embedding details.}
\notebaseline{}'s joint analysis-and-revision stage costs 215,633, 398,194,
284,637, and 311,598 tokens for DeepSeek/SWE, GPT-6 Luna/SWE,
DeepSeek/terminal, and GPT-6 Luna/terminal, respectively.
For ReasoningBank, the cache-key and retrieval-query
embedding subtotals are 6,265 and 9,220 on SWE-Chain-Evo with
\dsmodel{}, and 6,198 and 9,220 with \lunamodel{}. Terminal subtotals are 22,611 and 25,692 for \dsmodel{}, and 22,734 and 25,692 for \lunamodel{}.

\paragraph{Input and output.}
Across all settings, recorded input accounts for approximately 98.1\%
of the RSI Tax for both Meta-Harness and EvoSkill. These totals are consistent with repeated processing of growing
execution contexts, but input dominance alone does not identify
redundancy. Exact input/output attribution requires per-call usage;
matching character sequences can establish repeated inclusion of
trajectory text, but cannot determine token allocation within a prompt.

\section{Illustrative Update Traces}
\label{app:case-studies}
The following diagnostics examine proposal generation, acceptance
decisions, and memory retrieval in individual update traces. They use
different observation units and do not form an additive partition of
the RSI Tax. Observing one of these events does not establish that
removing its calls would preserve later performance.

\subsection{Event Criteria}
Table~\ref{tab:diagnostic-criteria} summarizes the criteria.

\begin{table}[!htbp]

\caption{\textbf{Supplementary event criteria for individual update traces.}
These diagnostics concern proposal generation, acceptance decisions,
and retrieval.}
\label{tab:diagnostic-criteria}

\centering\small
\begin{tabularx}{\linewidth}{@{}>{\raggedright\arraybackslash}p{0.23\linewidth}>{\raggedright\arraybackslash}X>{\raggedright\arraybackslash}p{0.25\linewidth}@{}}
\toprule
Pattern & Failure criterion & Unit and observation point \\
\midrule
Proposal-free search & A search round consumes tokens but emits no proposal. & Search round; after it ends. \\
\addlinespace
Decision-invariant evaluation & All possible validation results lead to the same acceptance decision. & Candidate-evaluation calls; condition known before evaluation. \\
\addlinespace
Unretrieved memory & A stored bundle has later task opportunities but is never retrieved before reset. & Bundle construction; retrieval observed through the remaining stream. \\
\bottomrule
\end{tabularx}
\end{table}

\subsubsection{Proposal-Free Search}
\label{sec:tax-search}\label{sec:pattern-proposal}
For a method that runs candidate search, let $C_u$ be the recorded cost
of round $u$, and let $q_u$ count
new proposals emitted by its proposer. A round is proposal-free when
$C_u>0$ and $q_u=0$. A proposal is counted before materialization or
validation, including one that later fails to build. Re-scoring an
existing candidate does not create a new proposal.

This pattern concerns failure to produce a proposed update, not the
absence of every possible benefit from search. A method may still
select a previously evaluated candidate when $q_u=0$, so deployment changes must
be recorded separately.

\subsubsection{Decision-Invariant Evaluation}
\label{sec:tax-evaluation}\label{sec:pattern-decision}
Fix the proposal, the acceptance rule, and all information available
before candidate evaluation. Let $A_{u,c}(v)\in\{0,1\}$ be the decision
to accept candidate $c$ at update $u$ if validation returns $v$. Let
$\mathcal V_{u,c}$ contain the validation outcomes not ruled out by that
pre-evaluation information. Evaluation is decision-invariant when
\begin{equation}
A_{u,c}(v)=A_{u,c}(v')
\quad\text{for all }v,v'\in\mathcal V_{u,c}.
\label{eq:decision-invariant}
\end{equation}
Thus, no possible result can change acceptance; an observed rejection
alone does not establish this condition. The recorded example is the
always-reject case: the incumbent scores 1, accuracy is bounded by 1,
and only a strict improvement on the same support set is accepted.
A rule that also rewards lower execution cost could still accept a
candidate at equal accuracy; it would require a different analysis.

The charged unit is candidate evaluation after this condition becomes
known. Incumbent evaluation used to establish the score, and analysis
or proposal generation in the same round, are not automatically part
of the diagnostic. Section~\ref{sec:result-decision} applies this test
to the strict selector used in our Meta-Harness implementation.

\subsubsection{Unretrieved Memory}
\label{sec:tax-reuse}\label{sec:pattern-memory}
For a retrieval-based memory bank, let $b_a$ be the judgment and
extraction cost of stored bundle $a$,
$h_a$ its remaining eligible task opportunities, and $r_a$ its retrieval
count before the next reset. Construction is classified as unretrieved
when $h_a>0$ and $r_a=0$, with total cost
\begin{equation}
N_{\rm unretrieved}=\sum_{a:h_a>0,\,r_a=0}b_a.
\label{eq:unretrieved}
\end{equation}
The unit is a bundle, which may contain several memory items. We exclude
updates that produce no bundle and construction after the final task.
Embedding tokens remain in total RSI usage but are not assigned to
bundle construction.

Retrieval records whether a bundle's contents entered the agent's task
context, not whether they were useful. A bundle can remain unretrieved because later
tasks do not need it; remaining task opportunities do not establish
semantic relevance. The criterion identifies a construction cost that
receives no retrieval during the observed horizon, rather than proving
the memory incorrect or permanently useless.

\FloatBarrier
\subsection{Case Studies}
\label{sec:patterns}\label{sec:phases}\label{sec:reuse}\label{sec:cases}
\begin{figure}[!t]
\centering
\includegraphics[width=\linewidth]{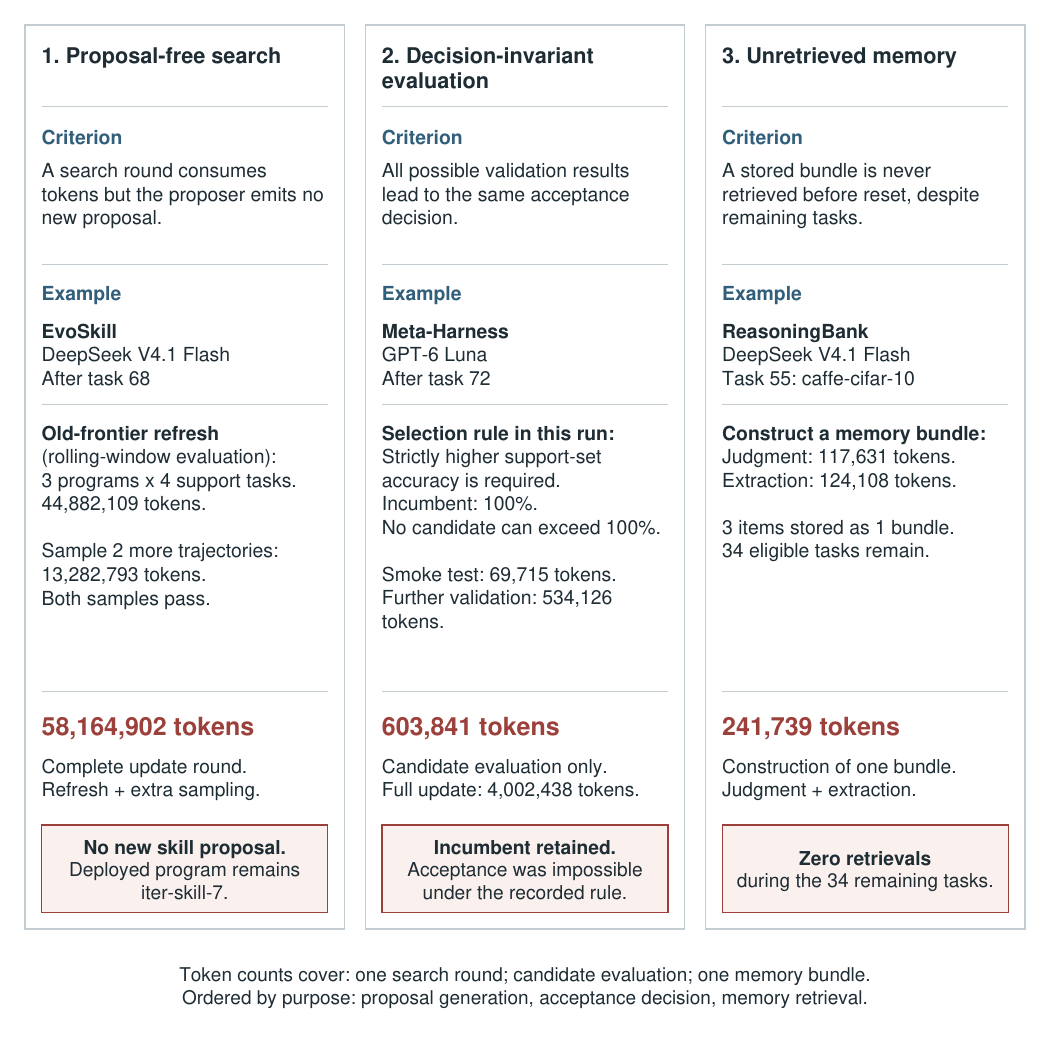}
\caption{\textbf{Illustrative update traces.}
Left: a search round without a proposal. Middle: candidate evaluation
whose result cannot change acceptance. Right: a stored memory bundle
with no later retrieval. The first two examples depend on the online
implementation choices described in Section~\ref{sec:environments}.}
\label{fig:diagnostics}
\end{figure}

We now apply the three criteria in their defined order
(Figure~\ref{fig:diagnostics}). Proposal-free search and decision-invariant
evaluation are illustrated by individual traces; the unretrieved-memory
diagnostic also has aggregate construction costs satisfying its predicate. Their
prevalence should not be inferred from the stage totals above.

\subsubsection{Proposal-Free Search}
\label{sec:result-proposal}
After task 68 in the DeepSeek EvoSkill run, three existing programs
are evaluated on four support tasks at a cost of 44,882,109 tokens.
The method then spends 13,282,793 tokens on two additional trajectories.
Both succeed, leaving no failed sample to trigger the proposer.
The complete round costs 58,164,902 tokens, produces no new proposal,
and retains \texttt{iter-skill-7}.

Most tokens in this round are spent re-evaluating the existing frontier
before a new proposal is available. The repeated refresh comes from the rolling
support window in our implementation; EvoSkill's original algorithm
uses a fixed validation set. The case motivates testing whether such
maintenance can be deferred or reduced when the proposal step has no
new change to offer.

\subsubsection{Decision-Invariant Evaluation}
\label{sec:result-decision}
After task 72 in the GPT-6 Luna Meta-Harness run, the incumbent scores
100\% on the support set and the selector requires a strictly higher
accuracy. Its acceptance function is therefore zero for every possible
candidate score. Nevertheless, the new candidate receives a 69,715-token
smoke test and 534,126 tokens of further validation.

The diagnostic covers the 603,841 candidate-evaluation tokens, rather
than all 4,002,438 tokens in the update. Incumbent evaluation may be
needed to establish the saturation condition. This example concerns
our strict deployment rule; the original Meta-Harness population search
does not require it. The identified calls cannot change the immediate
acceptance decision, although their traces may inform later search.

\subsubsection{Unretrieved Memory}
\label{sec:result-memory}
For task 55, \texttt{caffe-cifar-10}, in the DeepSeek ReasoningBank run,
judgment and extraction cost 117,631 and 124,108 tokens. They produce
one bundle of three items concerning the build and verification task.
None of the 34 remaining tasks retrieves that bundle. The bundle therefore contributes 241,739 construction tokens to
$N_{\rm unretrieved}$ in Equation~\ref{eq:unretrieved}.

This narrower unretrieved-bundle diagnostic excludes several costs
that the main accounting can classify as having no subsequent use.
Applying its predicate to all available memory records gives
53.1\%--61.5\% of ReasoningBank RSI tokens. On terminal tasks, the
corresponding construction costs are 5.16M of 8.61M RSI tokens for
DeepSeek and 3.62M of 5.88M for GPT-6 Luna. The numerator excludes
embeddings, updates without a stored bundle, and construction without
a later task opportunity; the denominator includes all RSI tokens.
These are cost shares, not fractions of memory items.
\FloatBarrier

%% file: rsi-impact-table.tex
\begin{table}[!htbp]
\centering\small\setlength{\tabcolsep}{3pt}
\begin{tabular*}{\linewidth}{@{\extracolsep{\fill}}lrrrrr@{}}
\multicolumn{6}{@{}l}{\textbf{(a) RSI costs by function: tokens (M) and share of RSI cost}} \\
\addlinespace[3pt]
\toprule
Method & RSI total & Verify & Explore & Write & Other \\
\midrule
Static & 0.00 & --- & --- & --- & --- \\
\addlinespace[3pt]
ReasoningBank & 48.98 & 0 & \shortstack[r]{24.61\\(50.25\%)} & \shortstack[r]{24.37\\(49.75\%)} & 0 \\
\addlinespace[3pt]
Meta-Harness & 1,648.86 & \shortstack[r]{1,233.89\\(74.83\%)} & \shortstack[r]{296.99\\(18.01\%)} & \shortstack[r]{91.74\\(5.56\%)} & \shortstack[r]{26.24\\(1.59\%)} \\
\addlinespace[3pt]
EvoSkill & 2,103.05 & \shortstack[r]{1,643.17\\(78.13\%)} & \shortstack[r]{455.29\\(21.65\%)} & \shortstack[r]{4.59\\(0.22\%)} & 0 \\
\addlinespace[3pt]
RRSI & 1,770.50 & \shortstack[r]{1,656.86\\(93.58\%)} & \shortstack[r]{45.22\\(2.55\%)} & \shortstack[r]{68.41\\(3.86\%)} & \shortstack[r]{0.009\\($<0.01\%$)} \\
\addlinespace[3pt]
\notebaseline{} & 1.21 & 0 & 0 & \shortstack[r]{1.21\\(100\%)} & 0 \\
\bottomrule
\end{tabular*}
\par\medskip
\begin{tabular*}{\linewidth}{@{\extracolsep{\fill}}llrrrrr@{}}
\multicolumn{7}{@{}l}{\textbf{(b) RSI costs by function and downstream impact (M)}} \\
\addlinespace[3pt]
\toprule
Method & Impact & Verify & Explore & Write & Other & Share (\%) \\
\midrule
ReasoningBank & With impact & 0.00 & 9.10 & 8.74 & 0.00 & 36.43 \\
 & No impact & 0.00 & 15.51 & 15.63 & 0.00 & 63.57 \\
\addlinespace[3pt]
Meta-Harness & With impact & 669.03 & 79.40 & 22.51 & 7.57 & 47.22 \\
 & No impact & 564.85 & 217.60 & 69.22 & 18.67 & 52.78 \\
\addlinespace[3pt]
EvoSkill & With impact & 1,025.71 & 276.29 & 3.75 & 0.00 & 62.09 \\
 & No impact & 617.46 & 179.01 & 0.84 & 0.00 & 37.91 \\
\addlinespace[3pt]
RRSI & With impact & 1,264.13 & 28.79 & 57.88 & 0.00 & 76.30 \\
 & No impact & 392.73 & 16.43 & 10.53 & 0.009 & 23.70 \\
\addlinespace[3pt]
\notebaseline{} & With impact & 0.00 & 0.00 & 1.21 & 0.00 & 100.00 \\
 & No impact & 0.00 & 0.00 & 0.00 & 0.00 & 0.00 \\
\bottomrule
\end{tabular*}
\caption{\textbf{Aggregate RSI costs by function and downstream impact.}
Both panels sum over the four model--environment settings. Panel (a) reports
function totals, with Other as a separate residual. Panel (b)
reports costs by function and impact; impact means integration
into subsequent task execution. Row shares use each method's complete
RSI cost, not the subtotal for an individual function. Cells,
function totals, and shares are independently rounded;
Static's function shares and impact status are undefined because its RSI cost is zero.}
\label{tab:impact-costs}
\end{table}